\documentclass[10pt,conference,a4paper]{IEEEtran}

\usepackage{times}
\usepackage{enumitem}

\usepackage[utf8]{inputenc}
\usepackage[english]{babel}
\usepackage{amsmath}
\usepackage{amsfonts}
\usepackage{amssymb}
\usepackage{graphicx}
\usepackage{bm} 

\usepackage[dvipsnames]{xcolor}

\usepackage{multirow, makecell}

\usepackage{caption}
\usepackage{subfig}

\usepackage[numbers]{natbib} 
\usepackage[hidelinks]{hyperref} 

\usepackage{wrapfig} 

\usepackage{float}
\floatstyle{plaintop}
\restylefloat{table}

\usepackage{natbib}

\usepackage{lineno}
\modulolinenumbers[5]

\title{Style-transfer GANs for bridging the domain gap in synthetic pose estimator training}
\author{\IEEEauthorblockN{Pavel Rojtberg} \IEEEauthorblockA{
Fraunhofer IGD, Darmstadt \\ TU Darmstadt \\ pavel.rojtberg@igd.fraunhofer.de}
\and
\IEEEauthorblockN{Thomas Pöllabauer}  \IEEEauthorblockA{
Fraunhofer IGD, Darmstadt \\ TU Darmstadt \\ thomas.poellabauer@igd.fraunhofer.de}
\and
\IEEEauthorblockN{Arjan Kuijper}  \IEEEauthorblockA{
Fraunhofer IGD, Darmstadt \\ TU Darmstadt \\ arjan.kuijper@igd.fraunhofer.de}
}

\makeatletter

\hypersetup{
pdftitle={\@title}
}

\begin{document}

\maketitle

\begin{abstract}
Given the dependency of current CNN architectures on a large training set, the
possibility of using synthetic data is alluring as it allows generating a virtually infinite amount of labeled training data.
However, producing such data is a non-trivial task as current CNN architectures are sensitive to the domain gap between real and synthetic data.

We propose to adopt general-purpose GAN models for pixel-level image translation, allowing to formulate the domain gap itself as a learning problem. The obtained models are then used either during training or inference to bridge the domain gap.
Here, we focus on training the single-stage YOLO6D \cite{tekin2018real} object pose estimator on synthetic CAD geometry only, where not even approximate surface information is available.
When employing paired GAN models, we use an edge-based intermediate domain and introduce different mappings to represent the unknown surface properties.

Our evaluation shows a considerable improvement in model performance when compared to a model trained with the same degree of domain randomization, while requiring only very little additional effort.
\end{abstract}

\IEEEpeerreviewmaketitle

\section{Introduction}
The ability to detect known objects and their 3D position relative to the viewer is crucial for Augmented Reality (AR) applications and many robotic tasks.
Especially AR applications require a precise estimation of distance and orientation, such that virtual content like annotations are correctly attached to real-world objects.

Recent advances with deep convolutional models such as SSD-6D \cite{kehl2017ssd}, PoseCNN \cite{Xiang-RSS-18} and YOLO6D \cite{tekin2018real} allow solving this problem using only RGB images in real-time.
However, to achieve state-of-the-art performance these models require a large amount of labeled training data. The assembly of such a training-set is an expensive, error-prone and time-consuming process \cite{hinterstoisser2019annotation}, making it cumbersome and often times inapplicable for use with custom applications.

When 3D geometry is available, one can resort to synthetically generate training data by rendering, which allows to create a virtually infinite training set in an automated fashion. However, it was shown that deep CNN models, even when applying cross-validation, tend to over-fit to the specific data-set \cite{torralba2011unbiased} and show significantly degraded performance when presented with data from a different domain \cite{ganin2016domain}.
Particularly, there is a strong \textit{domain-gap} between real and synthesized images, which typically prevents the use of synthetic images for training.

\begin{figure}
    \includegraphics[width=0.16\textwidth,height=0.15\textwidth]{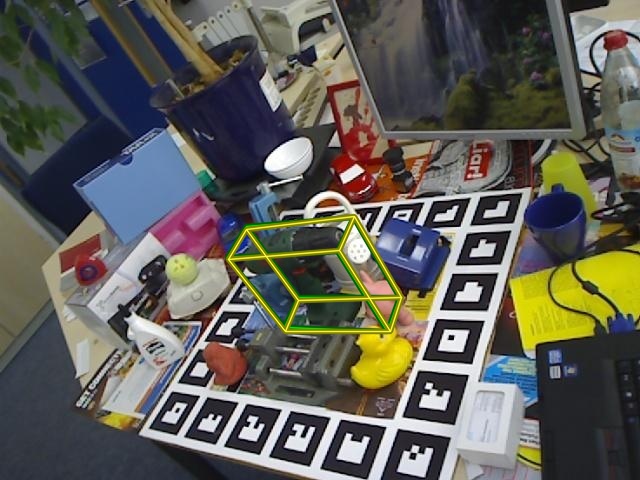}%
    \includegraphics[width=0.16\textwidth,height=0.15\textwidth]{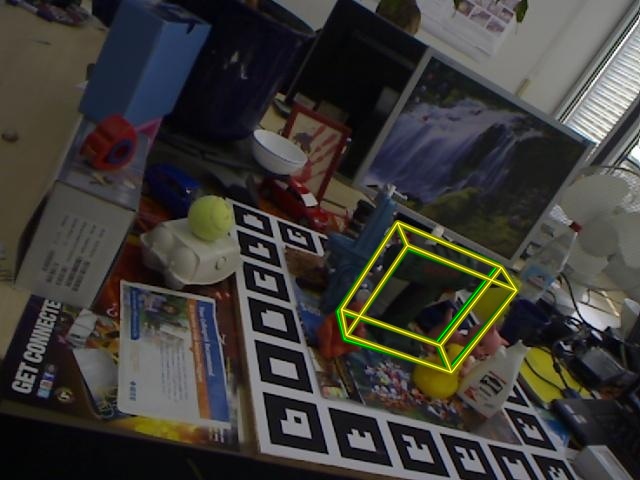}%
    \includegraphics[width=0.16\textwidth,height=0.15\textwidth]{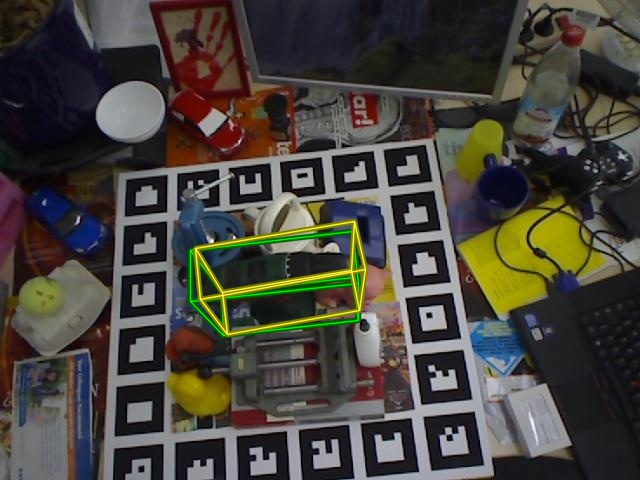}
\caption{Example of YOLO6D \cite{tekin2018real} on the \textsc{LineMod} data-set when trained \textit{solely} on geometry. The green bounding box represents the ground truth pose and the yellow one the predicted pose.}
\label{fig:cyclepred}
\end{figure}

To overcome this limitation, existing approaches apply \textit{domain randomization} (DR) to enforce domain invariance by overwhelming the model with variation \cite{tremblay2018training, tobin2017domain}, requiring the network to learn deeper, more abstract, features that are invariant across domains. An alternative direction is to reduce the gap by employing photo-realistic rendering \cite{tsirikoglou2017procedural} and structurally correct context generation \cite{prakash2019structured}.
Notably, \citet{tremblay2018deep} apply both for the task of object pose estimation.
However, designing a randomization method for a specific domain requires a domain expert to define which parts must stay invariant. Conversely, increasing photo-realism requires an artist to carefully model the specific environments in detail.
This in turn increases the cost of generating the data thus negating the primary selling point of using synthetic images in the first place.

When \textit{some} real images are available, transfer learning can be exploited, e.g. by fine-tuning a synthetically trained model with real images \cite{oquab2014learning}. Alternatively, \cite{hinterstoisser2018pre} propose to pre-train a network on real data and fine-tune on synthetic data.
Furthermore, it is possible to use the real data to enforce domain invariance during training \cite{ganin2016domain}. Recent work \cite{zakharov2019deceptionnet} extend this approach to guide domain-randomization, thus introducing some benefits of learning-based domain adaptation. However, it is still required to correctly select and design randomization modules.

Recent advances on generative adversarial networks (GANs) \cite{goodfellow2014generative, karras2018progressive, brock2018large} have shown great improvements regarding image quality and plausibility, training stability and variation of output.
Of particular interest in the task of closing the domain gap are conditional GANs \cite{isola2017image}. These are networks that, unlike traditional GANs, take additional inputs to condition generated output. Here, the image-conditional GANs form a general-purpose framework for image-to-image translation problems, like semantic segmentation, colorization and other style transfer tasks.
Existing solutions can be split into paired models \cite{isola2017image, wang2018high} and unpaired models \cite{CycleGAN2017}. The former are trained to adapt source to target images, paired in an supervised fashion, while the latter do not require supervision and instead directly learn to transfer the distribution of image features found in two unstructured data-sets.

This work focuses on employing such models to formulate the domain gap between real and synthetic images as a learning problem. At this, we propose training pipelines incorporating both paired and unpaired style-transfer and evaluate the results on the task of object pose estimation --- a particularly challenging scenario which requires a high fidelity of the object contours, which was, to the best of our knowledge, not previously addressed with GAN based domain adaptation.
In the context of paired style-transfer, we propose the use of the intermediate edge domain to do away with the need of real images for supervised training. Here, we evaluate different representation for mapping CAD geometry with unknown surface properties into the edge domain.

Most closely related to our work is \cite{antoniou2017data} which employ GANs for data augmentation. In contrast, we use GANs to specifically address the domain gap and employ synthetic data generation instead of data augmentation.
\cite{rambach2018learning} introduce the "pencil filter" as a domain with reduced expressiveness to tackle the domain gap and train a pose estimation network in this domain. However, the pose estimation network takes a strong performance hit as the "pencil domain" does not retain enough relevant features. We are able to avoid this hit by learning a mapping from the reduced domain to real images to reconstruct appropriate features.
In the medical domain, \cite{mahmood2018unsupervised} employ a GAN architecture for domain adaptation. However, they only consider the use of an unsupervised GAN model for reverse domain adaptation by making real images more synthetic, while we consider both paired and unpaired architectures and also consider the forward domain adaptation in the unsupervised case.

Based on the above, our key contributions are;
\begin{enumerate}[nolistsep]
\item formulating the domain gap as a learning problem using off-the-shelf image-conditional GANs,
\item introduction of the intermediate edge domain for training paired translation networks purely from synthetic data and
\item evaluation of paired and unpaired models regarding pose estimation performance.
\end{enumerate}

This paper is structured as follows: in Section \ref{sec:approach} the general approach is introduced and the choice of suitable GAN models is discussed.
In Section \ref{sec:evaluation} the method is evaluated in terms of pose estimation performance of the YOLO6D \cite{tekin2018real} model on the \textsc{LineMod} \cite{hinterstoisser2012model} data-set.

We conclude with Section \ref{sec:conclusion} giving a summary of our results and discussing the limitations and future work.

\section{Approach}
\label{sec:approach}
The core idea of our approach is to formulate the domain gap as a learning problem that is addressed with generative CNN models. Here, we use the generative adversarial framework to train a conditional generator that is able to augment images such that the pose estimation network becomes invariant to the source domain.
For this, the statistical distribution of image features found in both the real world and the synthetic domain must be matched, allowing the alignment of one domain to the other.

In this section we first discuss applicable GAN models and show qualitative results on the \textsc{LineMod} data-set to motivate the choice of specific image-conditional GAN models.
Then, we present our pipeline for fully synthetic training based on supervised image translation, leveraging \textsc{Pix2PixHD} \cite{wang2018high}.
Next, we turn to unsupervised image translation and introduce an alternative pipeline, that replaces the GAN model with \textsc{CycleGAN} \cite{CycleGAN2017}, which simplifies data acquisition by lifting the requirement of pairing images from both domains.

\begin{figure}
\subfloat[real texture] {
\includegraphics[width=0.24\textwidth]{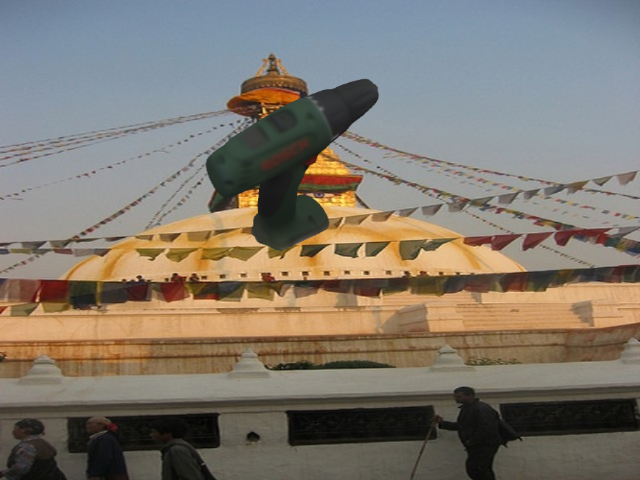}
}
\subfloat[random texture] {
\includegraphics[width=0.24\textwidth]{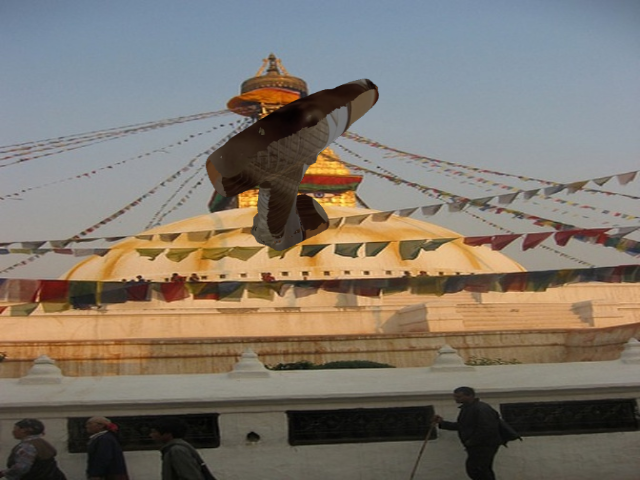}
}
\caption{Baseline data augmentation schemes for object "driller" using \textit{(a)} realistic object texturing and \textit{(b)} randomly selected image for object texturing}
\label{fig:baselines}
\end{figure}

\subsection{Baseline methods}
\label{sec:baseline}
We define two baseline methods for synthetic training. Both follow the data augmentation scheme of \cite{tekin2018real} by using random backgrounds from a set of real images, apply random image scaling and randomly adjust exposure and saturation adjustment. However, instead of using crops from real images, we render the object on top of the background (See Figure \ref{fig:baselines}) with the following methods:

\begin{enumerate}
\item[a)] Realistic texturing, by applying the true texture extracted from the data-set \cite{rojtberg2019real} and
\item[b)] randomly texturing the object during rendering.
\end{enumerate}

The first option depends on having the surface properties of the physical object available, but allows generating an arbitrary amount of realistic data by rendering. The second scheme applies blind DR by randomizing both the object and the background appearance. Note that this scheme neither requires plausible object placement nor plausible object appearance and thus is far easier to set up, compared to other DR solutions.

\subsection{Suitable GAN models}
The main requirement on the generator is that the resulting images have a high correlation with a given pose, such that the pose estimation model can be trained in a supervised fashion.
In theory any GAN model can be used for this if images can be mapped into the latent space of the generator. Such a mapping allows to condition the generator to create images, that resemble the input inside the \textit{latent space}.

If the latent space is constructed in a way that allows for interpolation by e.g. employing Kullback-Leibler loss \cite{diederik2014auto}, this approach would also allow synthesizing novel samples not seen during training.
In the case of pose estimation, this is required to generate images for views not seen by the adaptation network during training.
However, the resulting image must retain enough fidelity for the pose estimation network to predict the correct pose.
Not all GAN architectures are practical for this use case. For a preliminary experiment, we use the \textsc{StyleGAN} \cite{karras2019style} model, which offers state-of-the art generation performance and allows interpolation in the latent space of the generator.

\begin{figure}
\subfloat[input] {
\includegraphics[width=0.24\textwidth,height=0.18\textwidth]{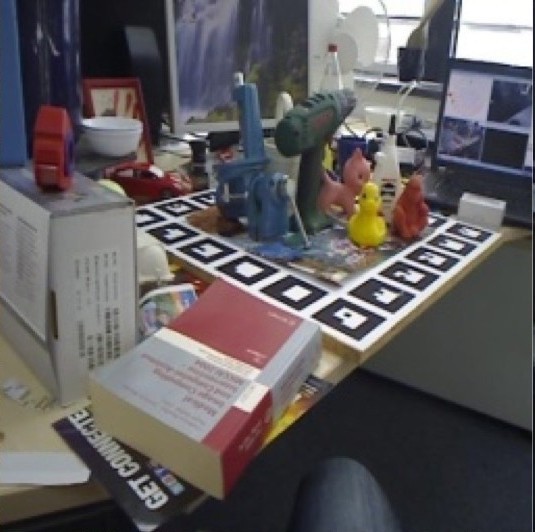}
}
\subfloat[sample] {
\includegraphics[width=0.24\textwidth,height=0.18\textwidth]{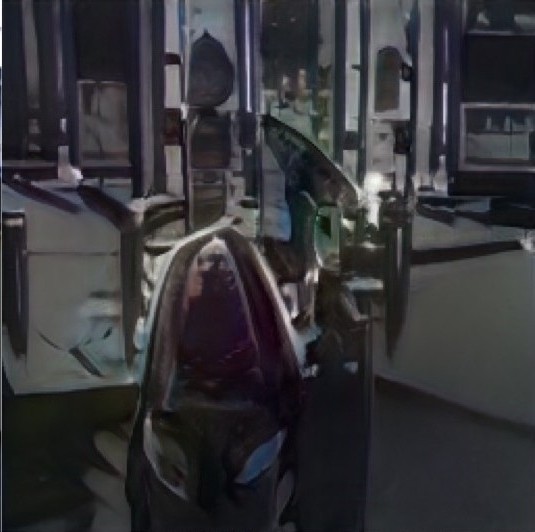}
}
\caption{Exemplary results of using \textsc{StyleGAN} for conditional sampling. Here, the input image is mapped into the latent space of the generator and is used as the mean for sampling.}
\label{fig:stylegan}
\end{figure}

Here, we map a real image into the latent space and use it as the mean for the random vector to sample new images, which would ideally retain most of the original image content; particularly the pose of the target object.
Figure \ref{fig:stylegan} shows an exemplary result; while the images seem plausible (considering the reduced training time), it is obvious that the model is not able to sufficiently capture the locality of the object. This results in a significantly different pose in the sampled image, compared to the input image. This precludes this approach in being used for pose estimation.
Therefore, we focus on conditional GAN models in the following.

For this experiment, \textsc{StyleGAN} was trained at a reduced resolution of $256 \times 256$ px for three days using approximately $110.000$ renderings composed on top of real backgrounds. The model was initialized using the weights for the LSUN Cat data-set \cite{yu2015lsun} of the original publication. This allowed operating with the reduced training period, compared to training the model from scratch which required 13 days according to the original publication.

\subsection{Paired intermediate domain translation}
\label{sec:paired}
In this section we introduce a training pipeline based on the \textsc{Pix2PixHD} \cite{wang2018high} paired image translation model. This is a supervised approach, which requires aligned image pairs to learn the domain translation. Pairing synthetic images with real images requires an according training set of real images, which defies the goal of synthetic training.
Therefore, we introduce a deterministic transform into an intermediate domain with reduced expressiveness, making real-world and rendered images less distinguishable.
Here, we use the Laplace filter, which approximates the second order image derivative that is continuous and directly translates to edge strength without requiring a thinning step.
The \textsc{Pix2PixHD} model is then employed to reconstruct real images from their Laplace filtered variants.

\begin{figure}
\includegraphics[width=0.49\textwidth]{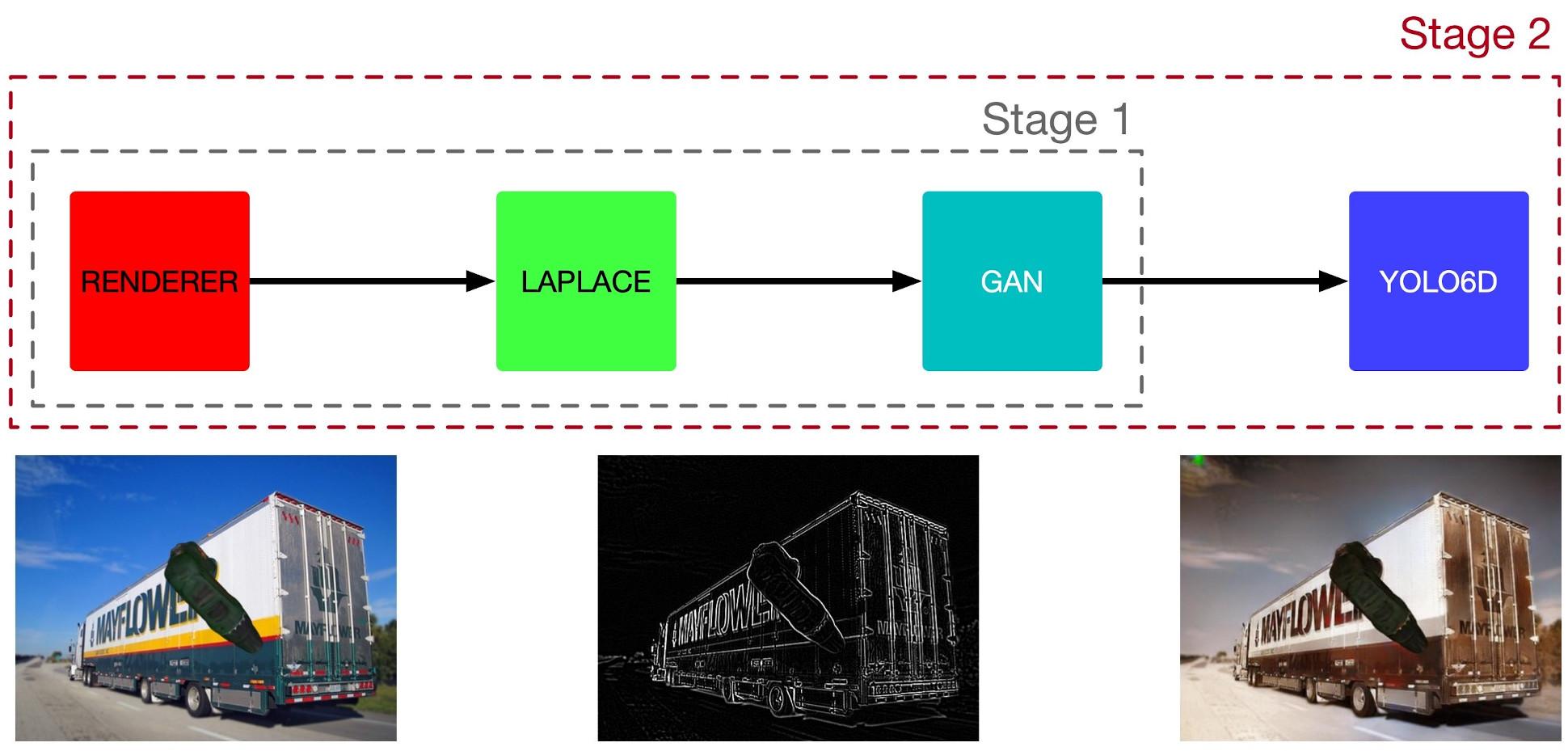}
\caption{Our synthetic training pipeline using the laplace filtered intermediate domain and incorporating the \textsc{Pix2PixHD} GAN for domain adaptation as presented in Section \ref{sec:paired}}
\label{fig:pix2pixpipe}
\end{figure}

The pipeline now consists of two trainable models; the \textsc{Pix2PixHD} model for domain adaptation followed by the pose-estimation task network (see Figure \ref{fig:pix2pixpipe}). As the performance of the second model depends on the first model, the pipeline has to be trained in two stages.
First, the domain adaptation network is trained until convergence on rendered images with random backgrounds and their Laplace-filtered variants. 
Here, the network must simultaneously reconstruct the rendering as well as the real background which forces the network towards realistic reconstructions. Next, the pose estimation model is trained on the reconstructed images.
Assuming the adaptation network was able to generalize from the training data, we now can create a virtually infinite amount of realistic views.

The remaining question is how to model the object surface before converting it to the Laplace domain, given that we do not want to impose any restrictions on possible object appearances. Here, one needs to balance the learning problem between the domain adaptation and the pose estimation network --- e.g. enforcing discriminative features makes the problem for the adaptation network more challenging, yet it reduces the difficulty for the pose estimation network.
Specifically, we opted for the following methods covering different work distributions of the involved networks:

\begin{figure}
\subfloat[random texture] {
\includegraphics[width=0.15\textwidth]{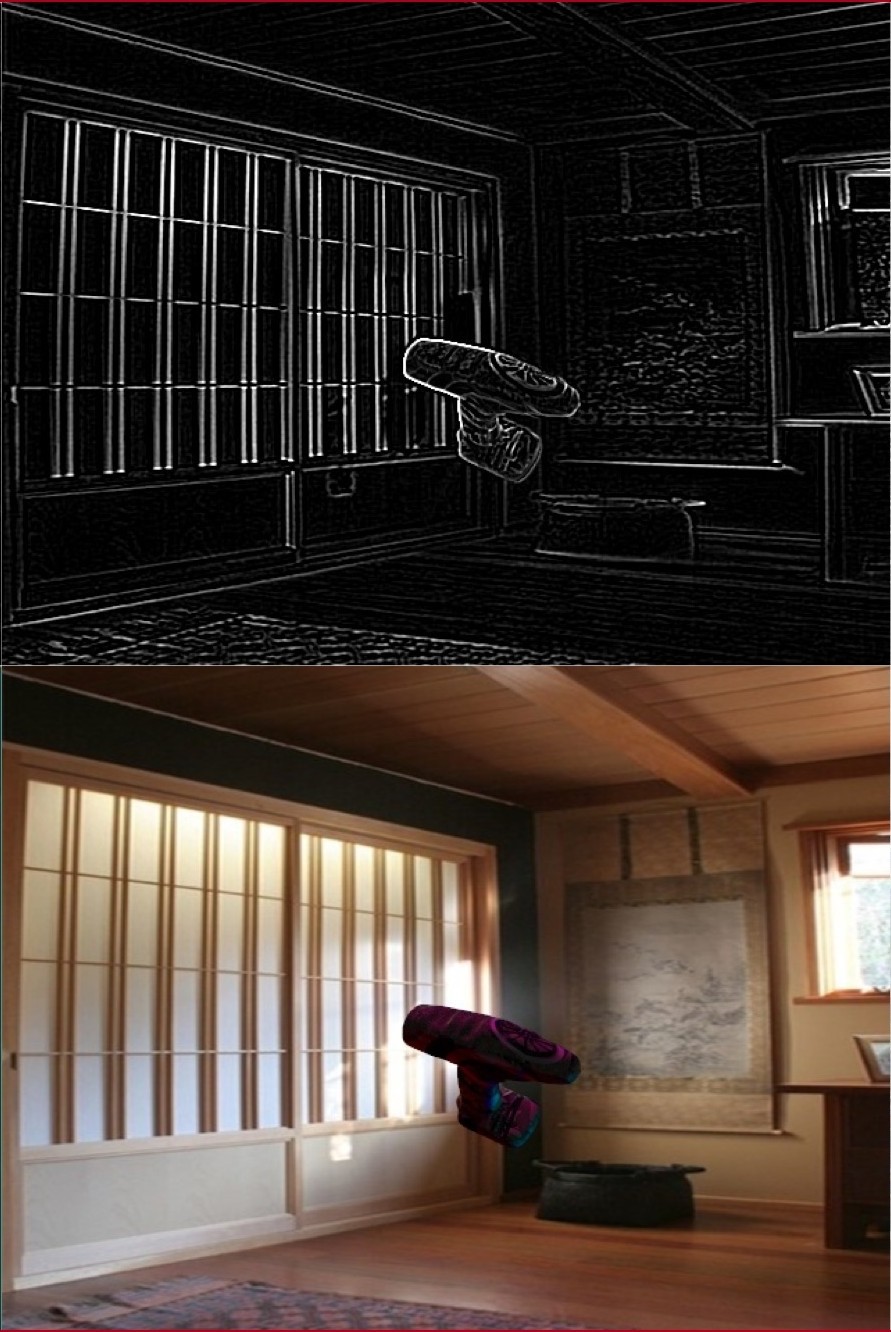}
\label{fig:pix2pix_random}
}
\subfloat[solid gray] {
\includegraphics[width=0.15\textwidth]{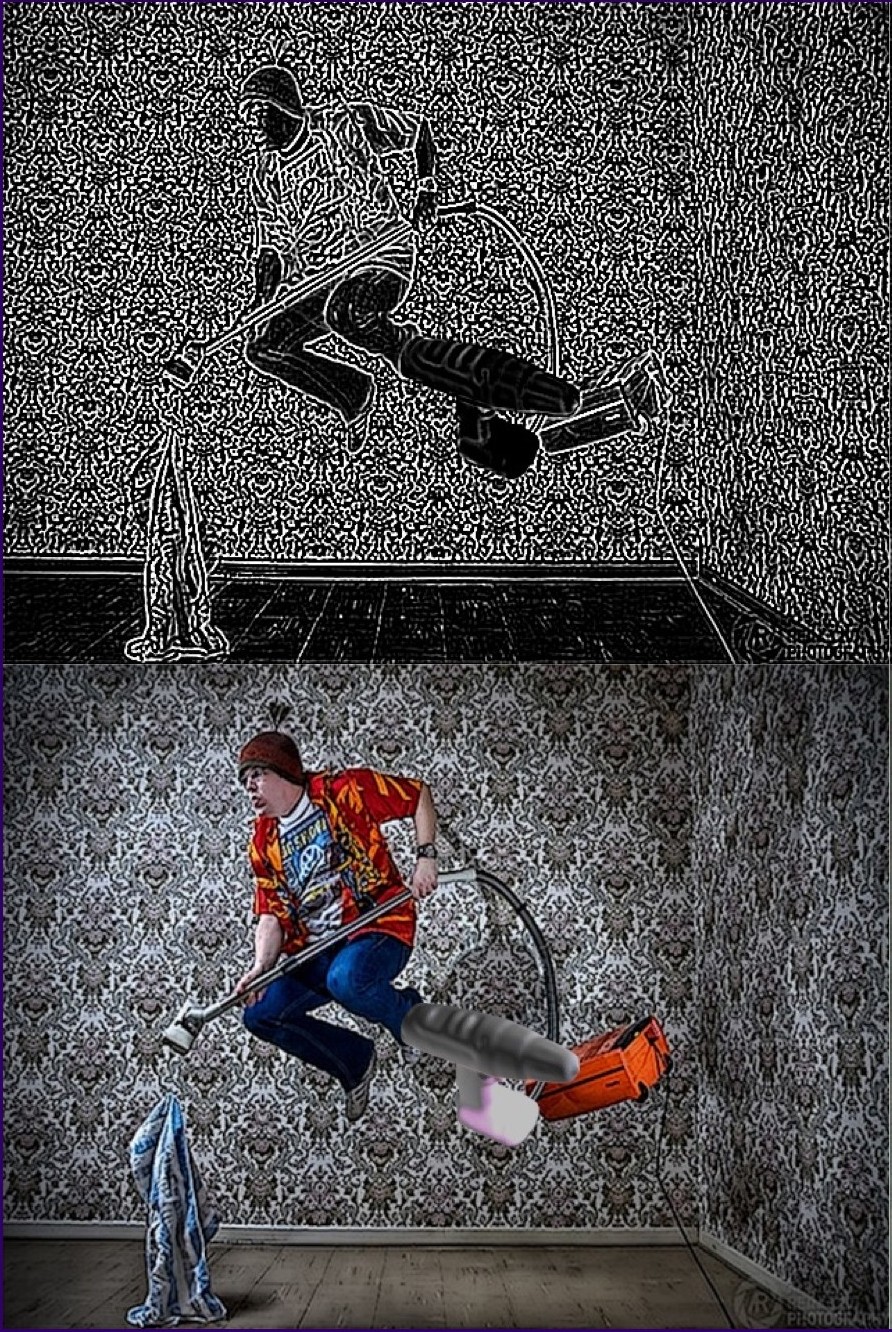}
\label{fig:pix2pix_gray}
}
\subfloat[checkerboard] {
\includegraphics[width=0.152\textwidth]{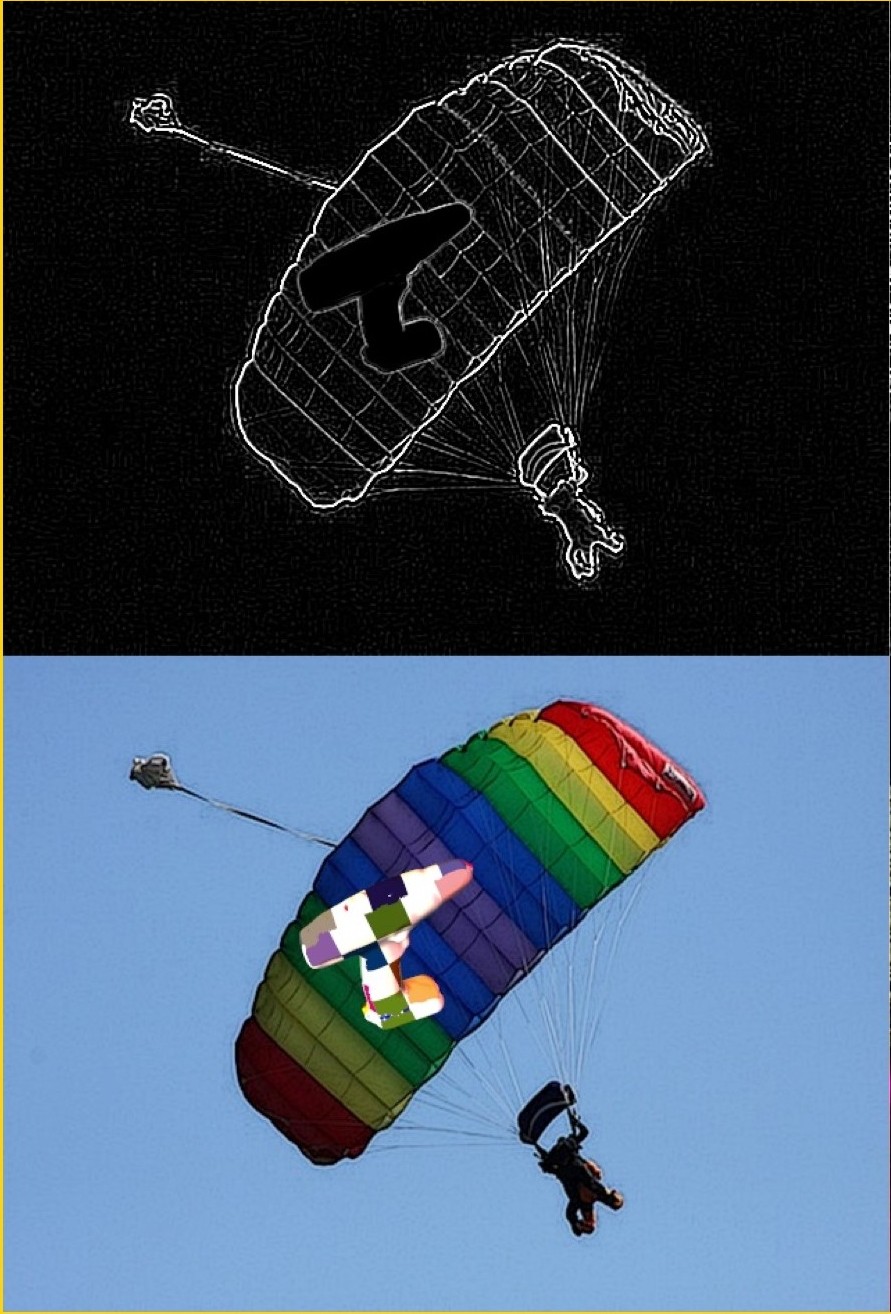}
\label{fig:pix2pix_checker}
}
\caption{Examples of the reconstruction tasks for \textsc{Pix2PixHD} with different rendering methods. \textit{Top row: }source image in the Laplace domain \textit{Bottom row:} target images to be reconstructed.}
\end{figure}

\begin{enumerate}
\item Use the real world texture (see Figure \ref{fig:pix2pixpipe}). This should result in the best model performance as it is the easiest task. However, to obtain the texture real images are needed, which dissents with our goal of fully synthetic training. This option was therefore mainly included to assess the performance loss of mapping into the Laplace domain as well as the loss induced by the following options. It corresponds to baseline variant a) with domain adaptation.

\item Use a random texture (see Figure \ref{fig:pix2pix_random}). This prevents the networks from learning any surface related information of the object to be detected. While this should not affect the domain adaptation network which is already faced with the task of reconstructing arbitrary background images, it makes the task of pose estimation significantly more challenging. The important shading and contour cues can be arbitrarily degraded by the used texture.
This corresponds to baseline variant b) with domain adaptation.

\item Use a uniform color for the object (see Figure \ref{fig:pix2pix_gray}). Instead of using a random texture, we assume the object to be of one uniform, yet arbitrary color. Given the Laplace intermediate domain, the adaptation network cannot learn a correct  object colorization. Therefore, we set the target color to gray, which is the most likely guess the adaptation network can take, given this task. Keeping the surface properties fixed allows to apply a consistent shading to the object, which in turn can be exploited by the pose estimation network.
However, the reconstruction of a plausible surface shading in turn makes the task of the domain adaptation network more challenging, but is a deliberate choice for balancing the work.
To prevent the adaptation network to over-fit to a specific lighting position, we place several point lights randomly around the object.

\item Use a fixed checkerboard pattern for the object surface (see Figure \ref{fig:pix2pix_checker}). Here, the Laplace images are generated from a uniformly colored object as above, while the reconstruction target is rendered with a fixed checkerboard texture. This makes the pose estimation task easier as the network can rely on stable cues on the object surface. However, this is particularly challenging for the adaptation network as it must encode the object geometry to reconstruct a correct appearance, while being confronted with merely an edge image generated from an uniformly colored object.
\end{enumerate}

Note that the translation of the object appearance to a different representation in the last two methods, requires the translation network to be executed at inference time as well.

\subsection{Direct image domain translation}
\label{sec:direct}
In this section we introduce a \textsc{CycleGAN} \cite{CycleGAN2017} based training pipeline for unsupervised domain adaptation. As this model does not require matching image pairs, there is no need for an explicit intermediate representation, and the model can directly learn the mapping between the domains of synthetic and real images.

\begin{figure}
    \centering
    \includegraphics[width=0.37\textwidth]{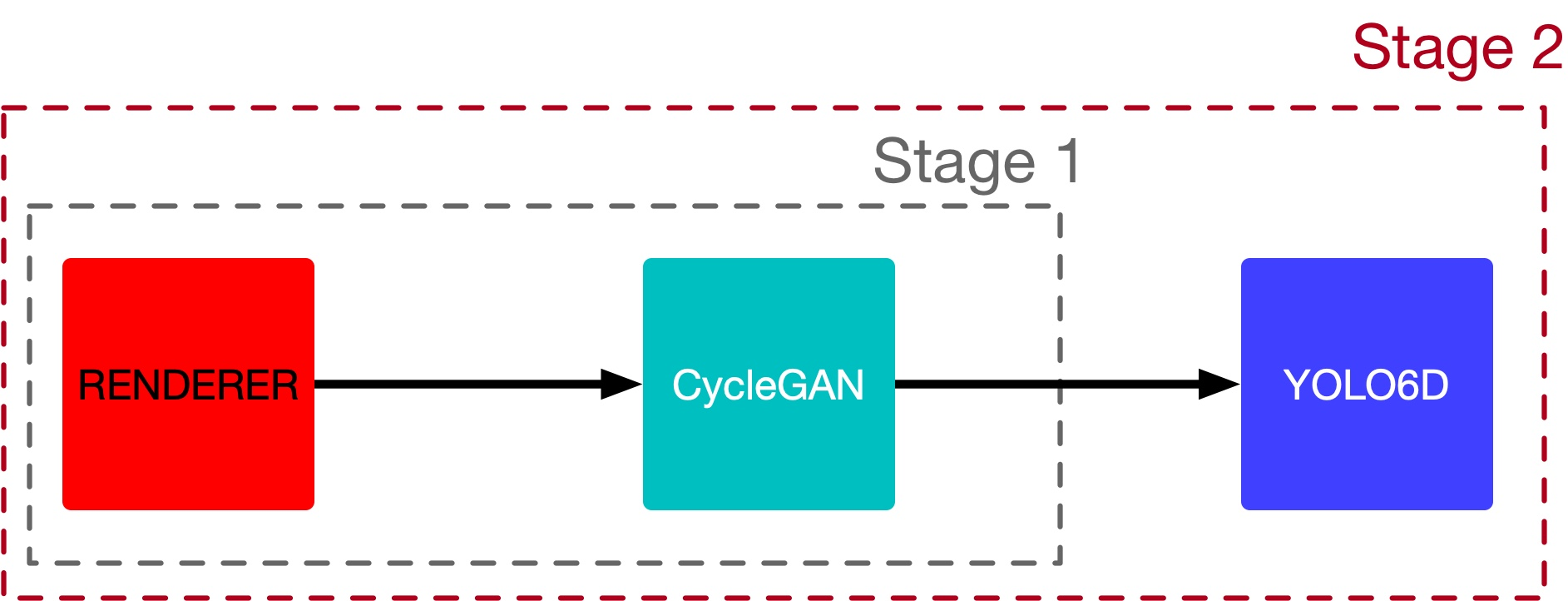}
    \caption{Our synthetic training pipeline using direct domain adaptation as presented in Section \ref{sec:direct}}
    \label{fig:cycleganpipe}
\end{figure}

\begin{figure}
    \subfloat[3D-game crops] {
    \includegraphics[width=0.23\textwidth]{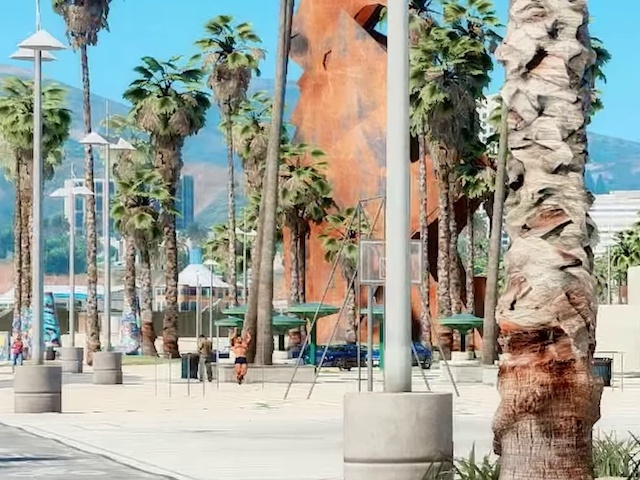}
    \includegraphics[width=0.23\textwidth]{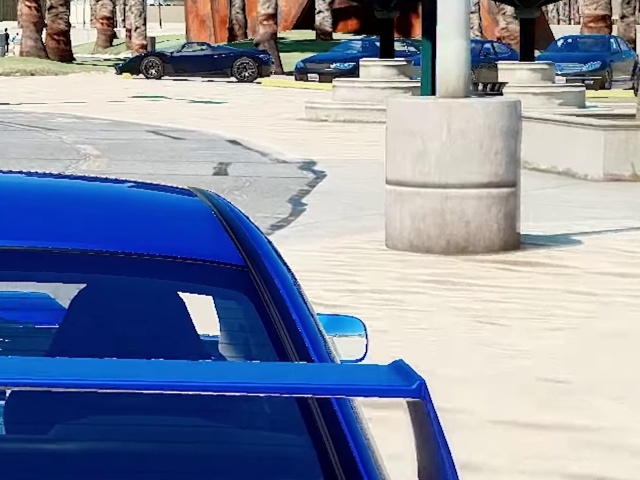}
    }
\\
    \subfloat[ImageNet] {
    \includegraphics[width=0.23\textwidth,height=0.17\textwidth]{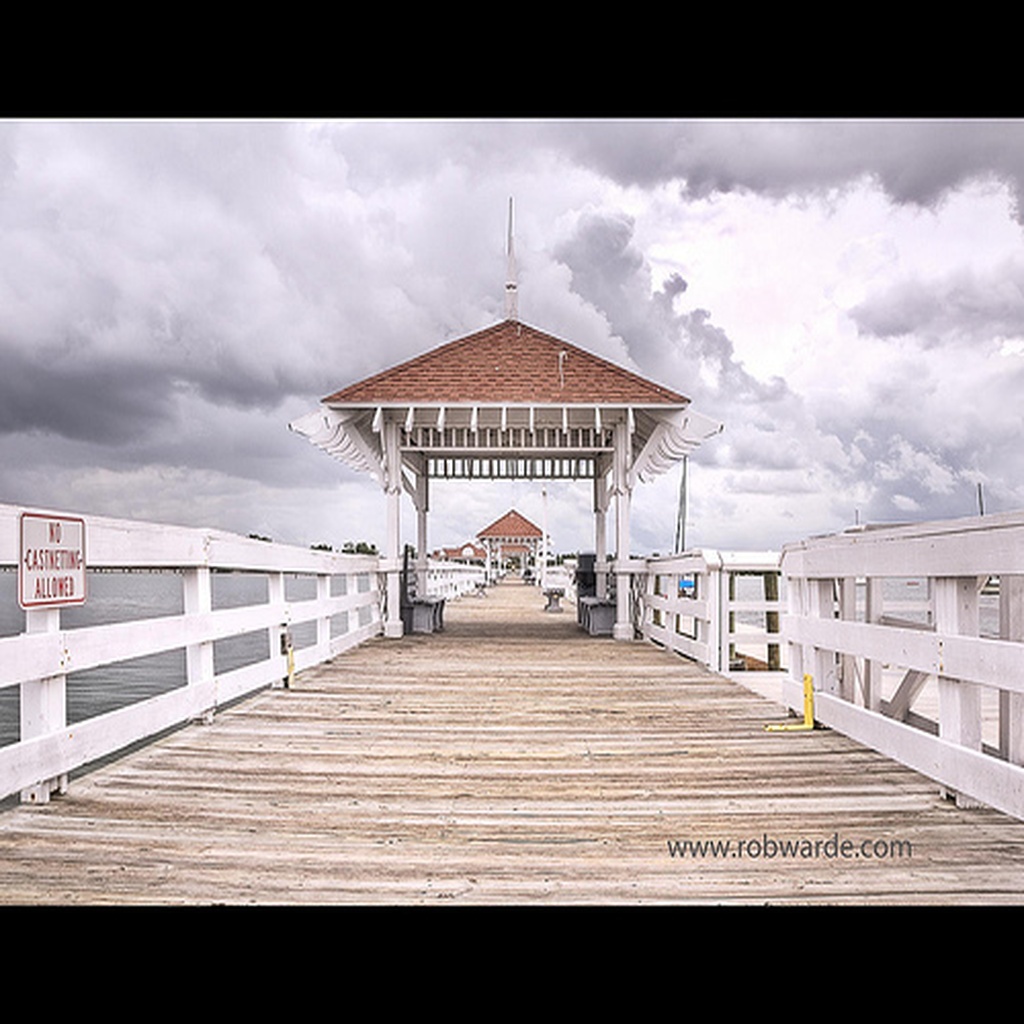}
    \includegraphics[width=0.23\textwidth,height=0.17\textwidth]{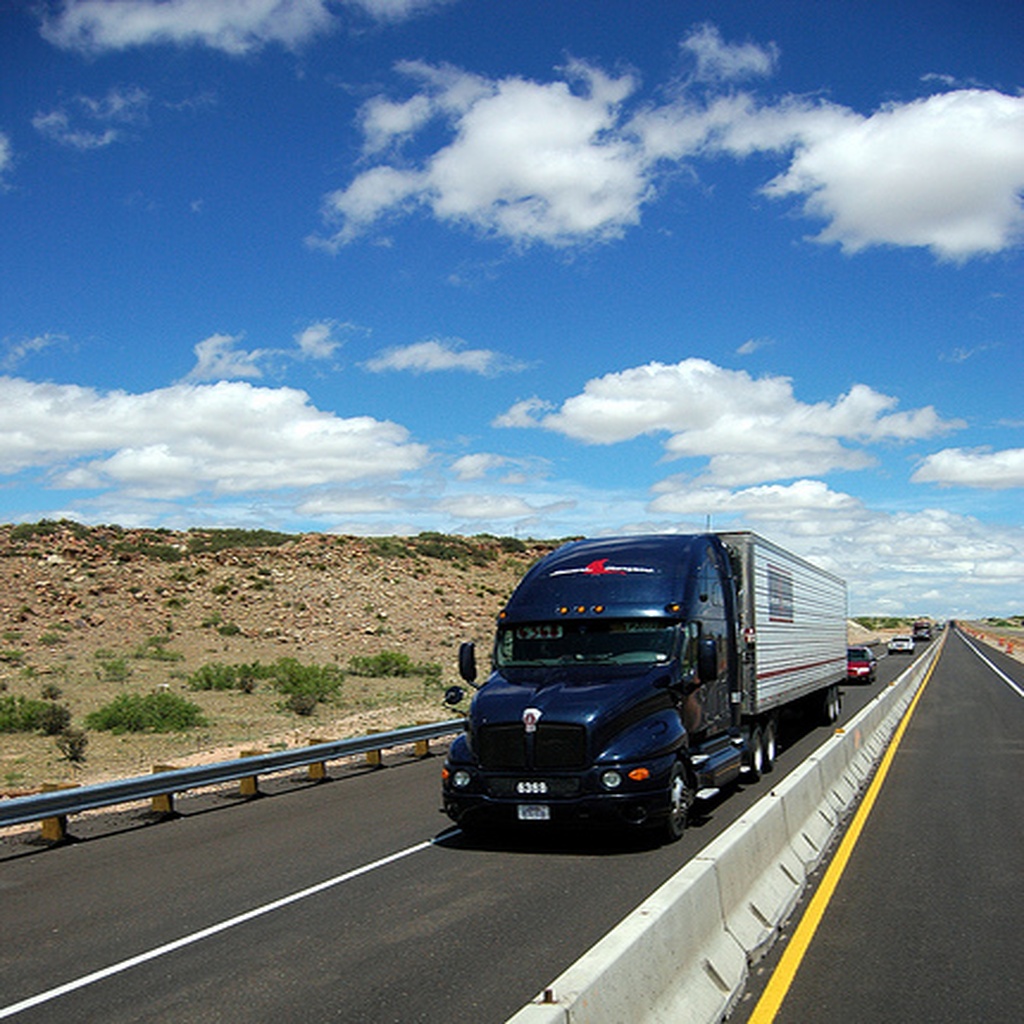}
    }
    \caption{Examples of the datasets, that we use as backgrounds and for training \textsc{CycleGAN}}
    \label{fig:3dgame}
\end{figure}

For the real domain we use the same generation scheme as in baseline b) of rendering randomly textured objects on random backgrounds. However, for the synthetic domain we cannot use real backgrounds to generate samples, as the different statistics would give away the synthetic object and bias the GAN towards object segmentation.
Instead, we collect a separate data-set of synthetic background images from 3D-game footage on Youtube. Here, we select ~$50.000$ random crops of randomly selected frames (see Figure \ref{fig:3dgame}) from a total of about 5 hours of video footage.

We train \textsc{CycleGAN} on the custom 3D-game and the \textsc{ImageNet} datsets to obtain a mapping between the real and the synthethic domain.
To train the YOLO6D detector, we generate synthetic renderings and then map them into the real domain with the learned \textsc{CycleGAN} model.
Furthermore, the \textsc{CycleGAN} architecture is capable of translating images in both directions. This allows reversing the pipeline; instead of adapting synthetic images to the real domain at training time, it is also possible to adapt real images to the synthetic domain at run-time.

As a limitation, the original \textsc{CycleGAN} model is tuned to produce images at a resolution of $256 \times 256$ px, which is only a fraction of the YOLO6D receptive field of $416 \times 416$ px. Scaling the GAN up would require fine-tuning its hyper-parameters like the size of the hidden layers, which is out of the scope of this work.
Therefore, we perform our experiments with the limited resolution, which allows to judge the feasibility of the method. However, one should keep in mind that pose precision can be improved by scaling the network output to match YOLO6D.

\section{Evaluation}
\label{sec:evaluation}

In this section we quantitatively and qualitatively evaluate, whether our pipeline allows the training of the demanding pose estimation task network from synthetic, randomized and non-photorealistic renderings only.
For comparability with related work, we use the \textsc{LineMod} data-set for evaluation. Here, we use the same test-train split as \cite{tekin2018real} for the following subset of objects: "ape", "can", "cam", "driller", "duck", "glue".

In the following, we first present the results of the baseline approaches as introduced in section \ref{sec:approach}. We then turn to the paired image translation via an intermediate, reduced domain and finally present the results for direct image domain translation.

\subsection{Implementation details}
For training the pipeline, we follow the procedure outlined by \cite{tekin2018real} in initially dropping the confidence loss, when training YOLO6D on a domain different from real images. This proved essential to allow the pose-estimator to adapt, as samples from the GAN exhibit significantly different colors and details than the \textsc{ImageNet} data-set, which the model was initialized on.
After $500.000$ samples the estimator was able to reach 85\% recall which improved to 95\% after $1.000.000$ samples.
Only after such initialization, we proceeded with training with the complete loss function.

Each benchmark used the following training parameters:
\begin{itemize}
\item stochastic gradient descent with momentum ($0.9$)
\item 2700 epochs
\item weight decay of $0.0005$ and learning rate of $0.001$
\item batch size of 32.
\end{itemize}
When not explicitly aiming for convergence, the learning rate was kept constant, otherwise it was reduced gradually with advancing training.

\subsection{Quantitative results}
We measure the domain translation performance of the presented methods in terms of pose estimation error of the task network \cite{tekin2018real}. Here, we employ two different metrics; the 3D translation and 3D rotation as well as the 2D corner re-projection error. The latter measures the error in screen-space and therefore is well suited for augmented-reality, while the former is more meaningful for robotic applications.

\begin{figure}
    \includegraphics[width=0.49\textwidth]{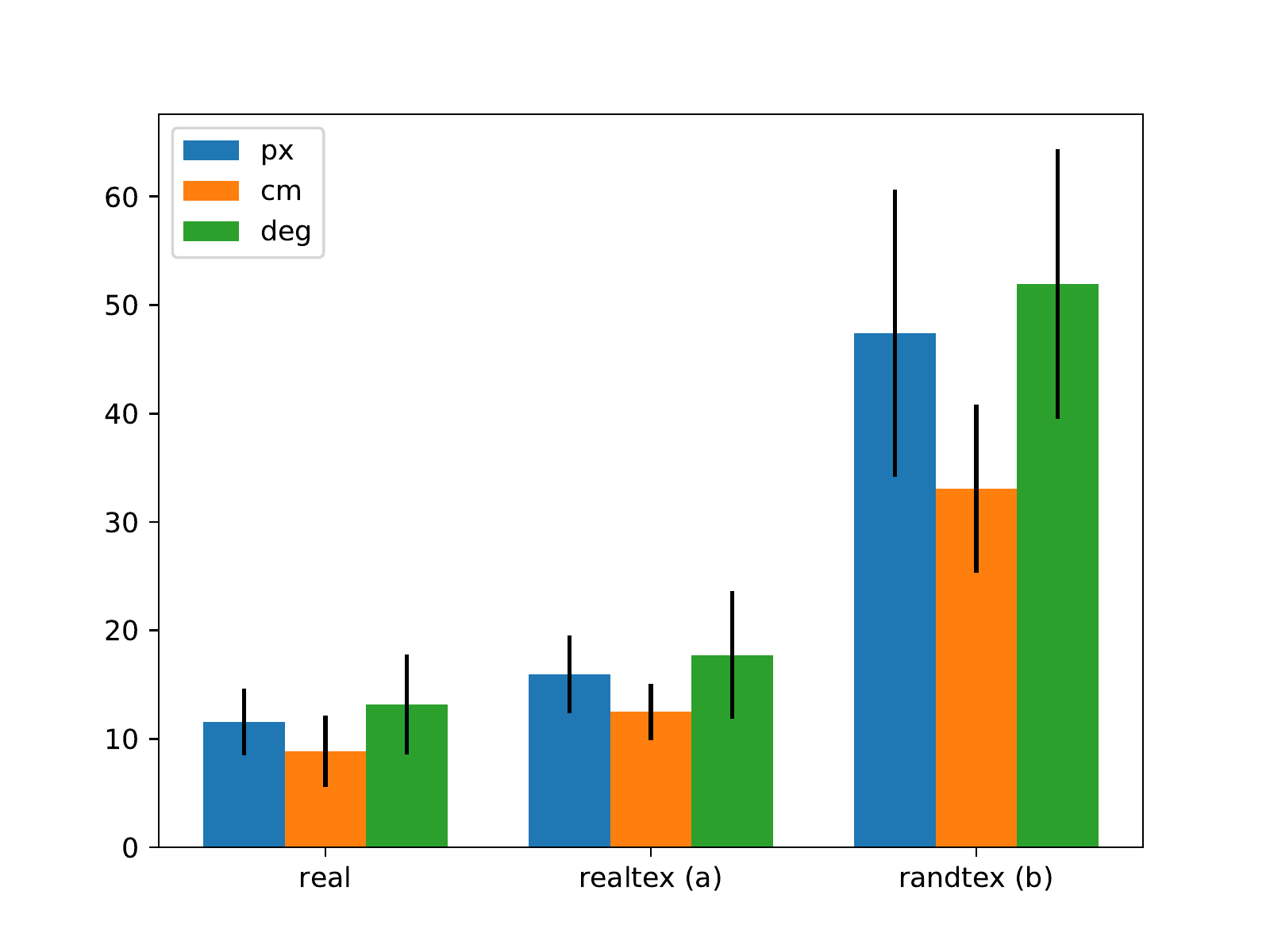}
    \caption{Pose estimation error of using real images as in \cite{tekin2018real} and the baseline methods a) and b) respectively.}
    \label{fig:evaluationBL}
\end{figure}

Comparing the baseline methods as introduced in Section \ref{sec:baseline} (see \autoref{fig:evaluationBL}), we see that the performance of baseline method b) is significantly reduced, although we ensured model convergence in both cases. This shows, that our task network for pose-estimation, YOLO6D, is not sufficiently conditioned to overcome the domain gap on its own, even when presented with a virtually unlimited amount of images from the synthetic domain.

\begin{figure}
    \includegraphics[width=0.49\textwidth]{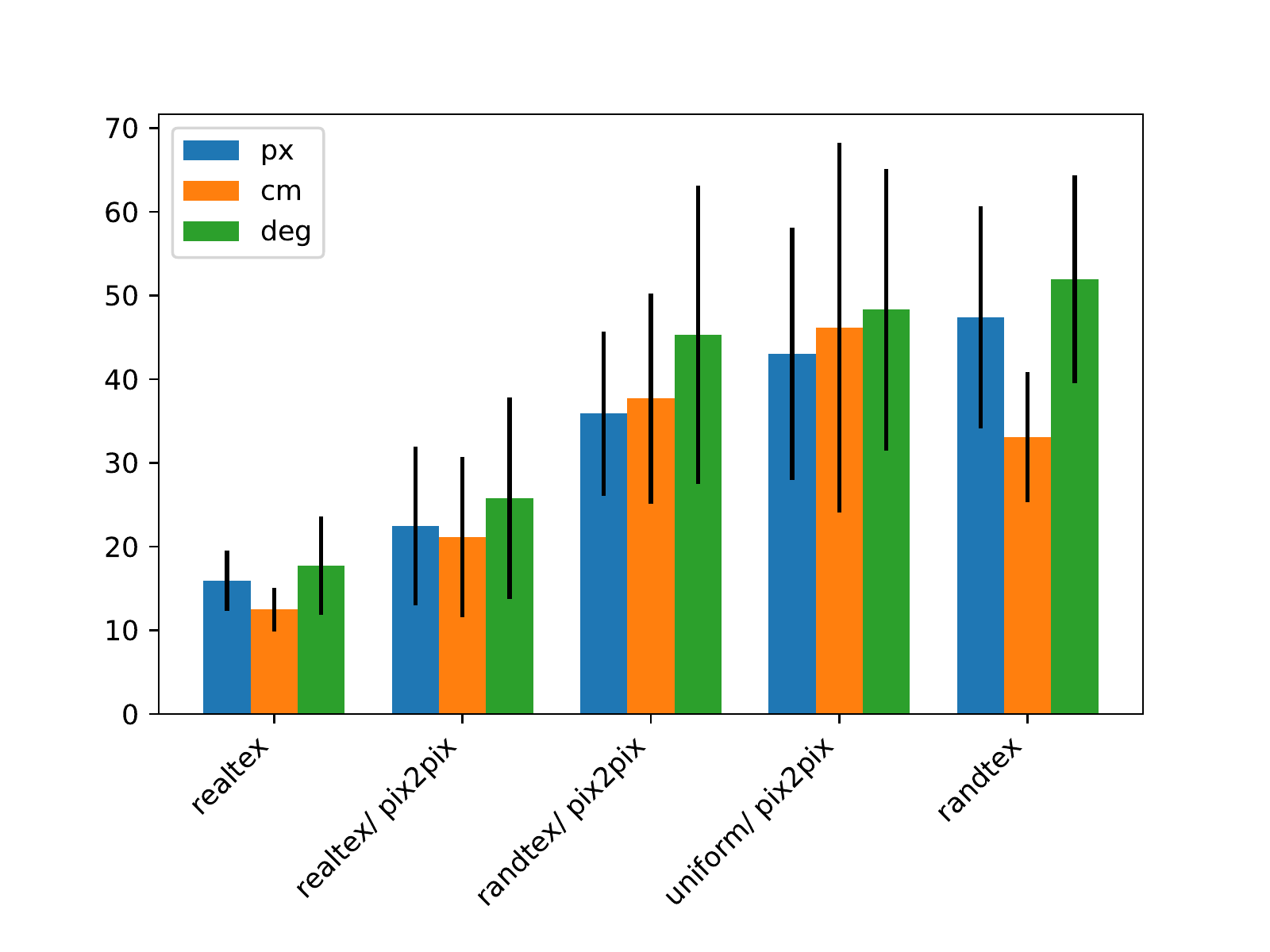}
    \caption{Pose estimation error of paired domain translation as introduced in Section \ref{sec:paired}}
    \label{fig:evaluationB}
\end{figure}

Looking at the domain translation results using the paired model as introduced in Section \ref{sec:paired} in \autoref{fig:evaluationB}, we see a reduction of re-projection error by about 23\% when using random texturing as in 2). This indicates that the edge-based intermediate representation leads to a more robust representation with YOLO6D. Likely, because we are able to reduce the texture bias \cite{geirhos2018imagenettrained} of the model.
In contrast, rendering the object using a solid color as in 3) results in degraded results compared to using random texturing. Probably the pose-estimator over-fits to the uniform colour present in the reconstruction, thus still suffering from the domain-gap.
Training the pose estimation network was not possible when applying rendering-method 4).

While the results improve over baseline, the margin is only moderate. Likely, the reason for this is that the Laplace filtering does not sufficiently reduce the expressiveness of the image and the domain gap is still present in the intermediate representation.

\begin{figure}
    \includegraphics[width=0.49\textwidth]{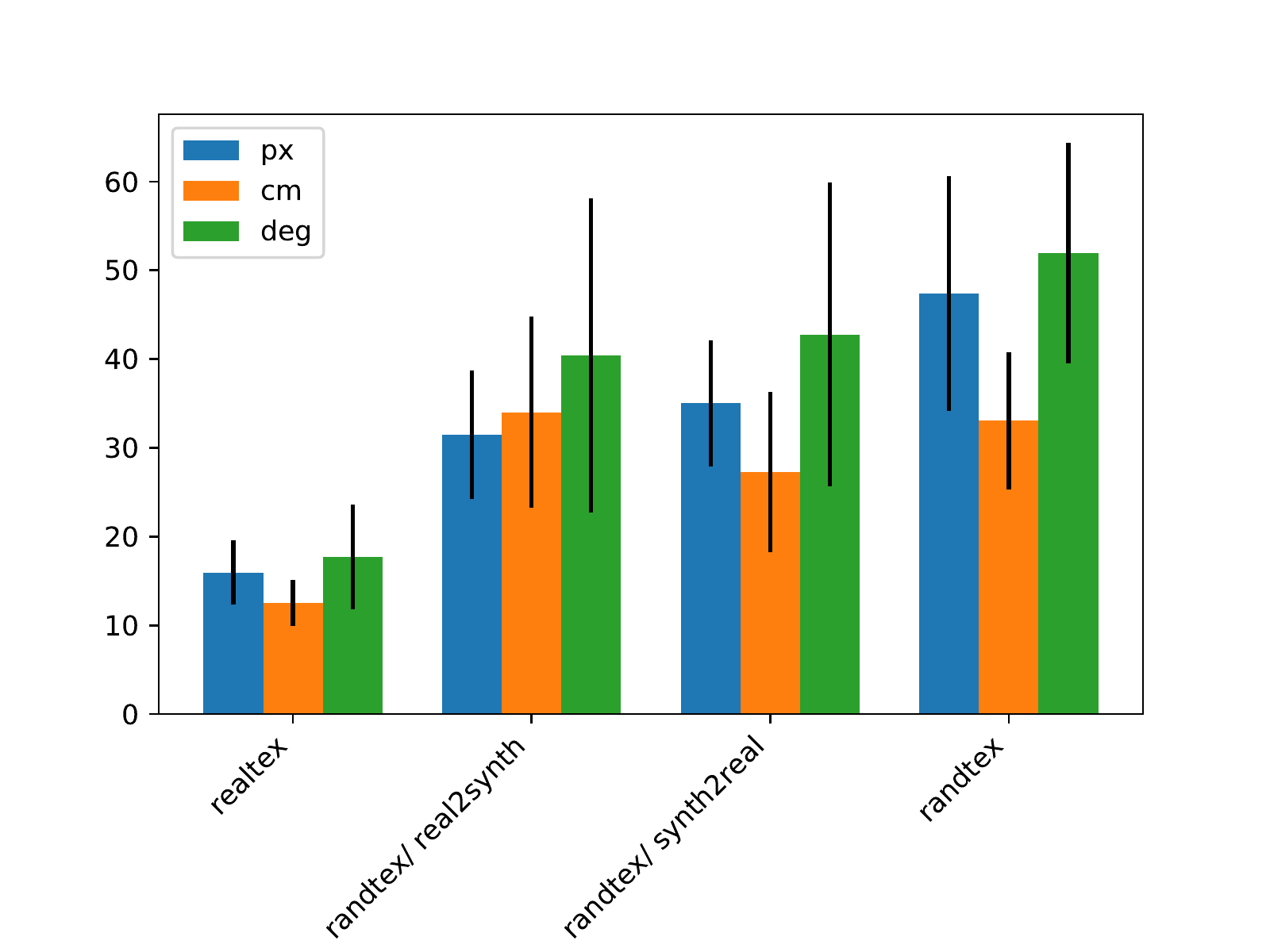}
    \caption{Pose estimation error of direct domain translation as introduced in Section \ref{sec:direct}, as well as translation reversal (real2synth).}
    \label{fig:evaluationC}
\end{figure}

Turning to the direct domain translation using \textsc{CycleGAN} as introduced in Section \ref{sec:direct} in \autoref{fig:evaluationC}, the results of synthetic to real translation are consistent with paired translation, even though the network produces images only at quarter the resolution. This leads us to believe that unsupervised models are generally sufficient for domain adaptation.

When reversing the translation from real to synthetic, we see a reduction of re-projection error by 31\% compared to random texturing (b). However, the real-textured baseline (a) has a 50\% lower error.

\begin{table}
    \centering
    \begin{tabular}{|c|c|c|c|}
        \hline 
        \multirowcell{2}{Rendering method /\\Domain translation} & \multicolumn{3}{c|}{Mean error} \\
        \cline{2-4} & re-projection & translation & angle\\ 
        \hline
        \hline
        real images \cite{tekin2018real} & 12 px & 8.9 cm & $13.2^{\circ}$ \\
        \hline
        realtex / none & 16 px  & 12.5 cm &  $17.7^{\circ}$  \\ 
        \hline 
        randtex / none & 47 px   & 33 cm & $52.0^{\circ}$ \\ 
        \hline
        realtex / laplace & 22.5 px  & 21.2 cm & $25.8^{\circ}$ \\
        \hline
        randtex / laplace & 36 px & 37.6 cm & $45.3^{\circ}$ \\
        \hline
        uniform / laplace & 43 px & 46.1 cm & $48.3^{\circ}$  \\
        \hline
        randtex / real2synth & 35 px & 27.2 cm & $42.8^{\circ}$ \\
        \hline
        randtex / synth2real & 32 px  & 34 cm & $40.4^{\circ}$ \\
        \hline 
    \end{tabular} 
    \caption{Raw data for \autoref{fig:evaluationBL}, \autoref{fig:evaluationB}, \autoref{fig:evaluationC}}
\end{table}

\subsection{Qualitative results}

\begin{figure}
\includegraphics[width=0.49\textwidth]{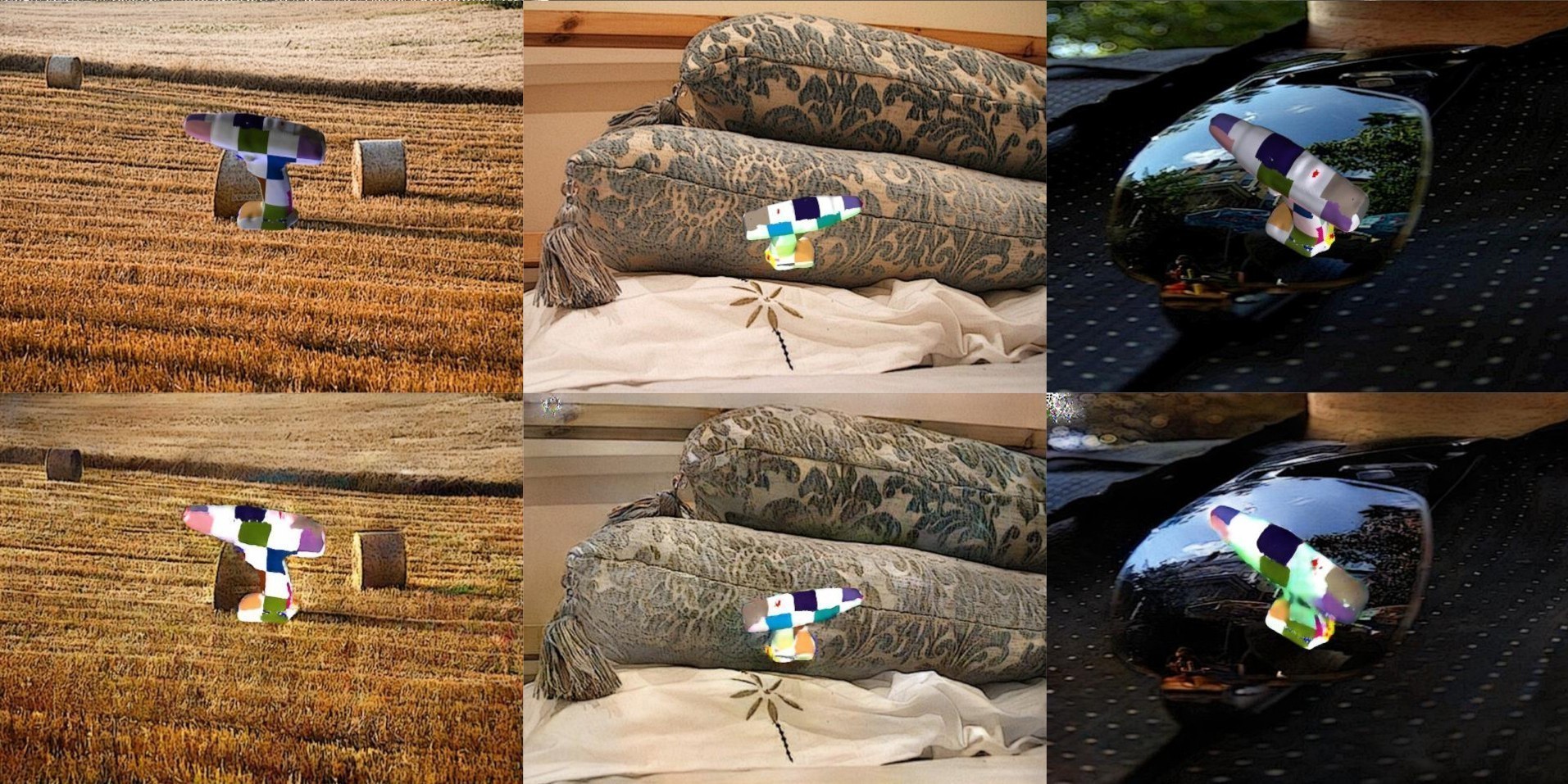}
\caption{Checkerboard surface reconstruction results as required by surface rendering method 4). \textit{Top row:} target image. The input to the network is a Laplace filtered version (confer Figure \ref{fig:pix2pix_checker}) \textit{Bottom row:} output of the \textsc{Pix2PixHD} model trained to reconstruct the checkerboard pattern and applied to the whole image.}
\label{fig:pix2pix_checker_re}
\end{figure}

As shown in \autoref{fig:pix2pix_checker_re}, \textsc{Pix2PixHD} can not predict a consistent checkerboard texture. While the results are plausible, they do not exhibit enough detail to help the pose estimation model.
Still, this is a remarkable result as the translation network is confronted with a significantly more demanding challenge; to correctly apply the texture pattern it not only has to infer the object pose, but also the object geometry.

Figure \ref{fig:cyclepred} shows some predictions of the pipeline trained with \textsc{CycleGAN} for domain adaptation. While the model is able to reliably detect the object, the accuracy of the object pose is lacking. This is likely due to the limited resolution of the adaptation network and can be further improved by scaling it up to the pose estimation input size.

\begin{figure}
\includegraphics[width=0.49\textwidth]{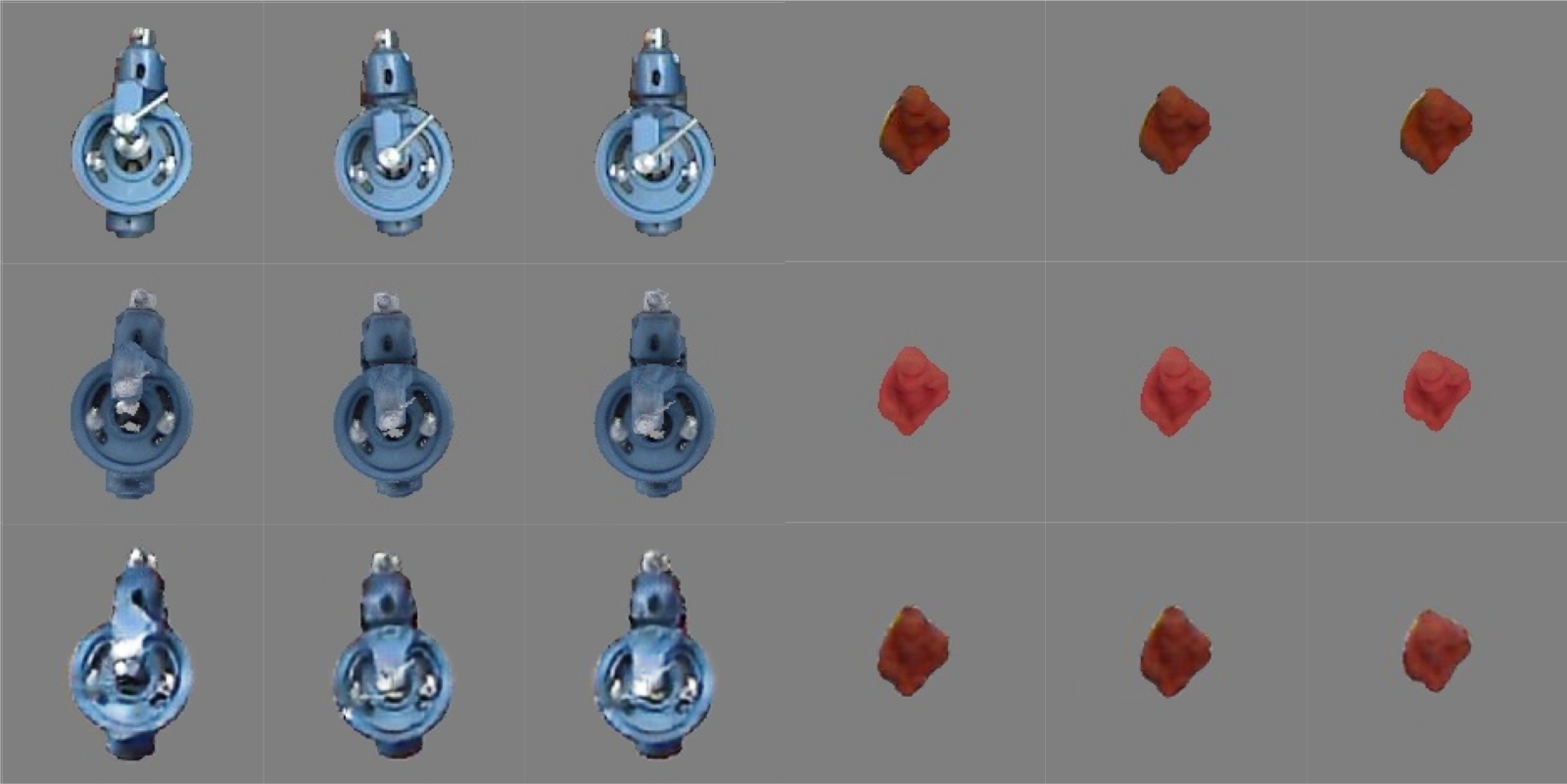}
\caption{Domain translation capabilites of \textsc{CycleGAN}. \textit{Top row:} masked crops from real images. \textit{Middle row:} synthetic rendering without lighting and using vertex colors only. \textit{Bottom row:} output of \textsc{CycleGAN} applied on the middle row, adding lighting effects and generally improving image fidelity.}
\label{fig:cycle_qual}
\end{figure}

Figure \ref{fig:cycle_qual} shows the effect on image quality after applying \textsc{CycleGAN} on the objects "benchvise" and "ape".  The model is able to improve color reproduction as well as applying details like specular highlights.

\section{Conclusion}
\label{sec:conclusion}

We have shown that employing a paired translation GAN for domain adaptation during training generally improves model robustness and hence the performance of the target network. Turning to unpaired translation GANs, we have shown that training solely on CAD geometry with neither knowing the surface properties nor the environment is possible and is in the same error range as using paired GANs --- with a slight advantage, when using real2synth translation.
These results indicate that image-conditional GANs are indeed an effective measure to close the domain gap between real and synthetic images.
However, when only relying on the CAD geometry with both the object background and object surface being randomized, there is still a considerable performance degradation.

On the other hand, the introduced training is much simpler compared to existing solutions relying on domain randomization. The latter require a faithful setup of randomization modules --- even when employing guided domain randomization.
In contrast, the presented style transfer pipelines only require collecting unstructured images from the target domain, which are fed into the adaptation network in an unsupervised fashion. This allows focusing on training the task network. At this, we have shown that it is possible to train a pose-estimation model with satisfactory precision.

To further improve the performance of our approach, the obvious measure is to take out the resolution loss by scaling up the \textsc{CycleGAN} output to match the YOLO6D input. 
Furthermore, the method currently relies on blind randomization during rendering to consider the multitude of possible surface materials. Instead, it seems beneficial to employ a generator that is aware of the multi-modal data and therefore is capable to produce different surface materials based on a noise vector.
To this end, one could replace the image-conditional \textsc{CycleGAN} by a more recent variant like MUNIT \cite{huang2018multimodal}. Alternatively, further regularization on the unconditional \textsc{StyleGAN} \cite{karras2019style}, could enforce its samples closely resemble the input image.
In this case one could reconsider the architecture as it is also capable to generate multiple styles, by internally disentangling content and style.

\bibliographystyle{abbrvnat}
\bibliography{bibliography}

\begin{thebibliography}{30}
\providecommand{\natexlab}[1]{#1}
\providecommand{\url}[1]{\texttt{#1}}
\expandafter\ifx\csname urlstyle\endcsname\relax
  \providecommand{\doi}[1]{doi: #1}\else
  \providecommand{\doi}{doi: \begingroup \urlstyle{rm}\Url}\fi

\bibitem[Antoniou et~al.(2017)Antoniou, Storkey, and Edwards]{antoniou2017data}
A.~Antoniou, A.~Storkey, and H.~Edwards.
\newblock Data augmentation generative adversarial networks.
\newblock \emph{arXiv preprint arXiv:1711.04340}, 2017.

\bibitem[Brock et~al.(2019)Brock, Donahue, and Simonyan]{brock2018large}
A.~Brock, J.~Donahue, and K.~Simonyan.
\newblock Large scale {GAN} training for high fidelity natural image synthesis.
\newblock In \emph{International Conference on Learning Representations}, 2019.

\bibitem[Diederik et~al.(2014)Diederik, Welling, et~al.]{diederik2014auto}
P.~K. Diederik, M.~Welling, et~al.
\newblock Auto-encoding variational bayes.
\newblock In \emph{Proceedings of the International Conference on Learning
  Representations (ICLR)}, volume~1, 2014.

\bibitem[Ganin et~al.(2016)Ganin, Ustinova, Ajakan, Germain, Larochelle,
  Laviolette, Marchand, and Lempitsky]{ganin2016domain}
Y.~Ganin, E.~Ustinova, H.~Ajakan, P.~Germain, H.~Larochelle, F.~Laviolette,
  M.~Marchand, and V.~Lempitsky.
\newblock Domain-adversarial training of neural networks.
\newblock \emph{The Journal of Machine Learning Research}, 17\penalty0
  (1):\penalty0 2096--2030, 2016.

\bibitem[Geirhos et~al.(2019)Geirhos, Rubisch, Michaelis, Bethge, Wichmann, and
  Brendel]{geirhos2018imagenettrained}
R.~Geirhos, P.~Rubisch, C.~Michaelis, M.~Bethge, F.~A. Wichmann, and
  W.~Brendel.
\newblock Imagenet-trained {CNN}s are biased towards texture; increasing shape
  bias improves accuracy and robustness.
\newblock In \emph{International Conference on Learning Representations}, 2019.

\bibitem[Goodfellow et~al.(2014)Goodfellow, Pouget-Abadie, Mirza, Xu,
  Warde-Farley, Ozair, Courville, and Bengio]{goodfellow2014generative}
I.~Goodfellow, J.~Pouget-Abadie, M.~Mirza, B.~Xu, D.~Warde-Farley, S.~Ozair,
  A.~Courville, and Y.~Bengio.
\newblock Generative adversarial nets.
\newblock In \emph{Advances in neural information processing systems}, pages
  2672--2680, 2014.

\bibitem[Hinterstoisser et~al.(2012)Hinterstoisser, Lepetit, Ilic, Holzer,
  Bradski, Konolige, and Navab]{hinterstoisser2012model}
S.~Hinterstoisser, V.~Lepetit, S.~Ilic, S.~Holzer, G.~Bradski, K.~Konolige, and
  N.~Navab.
\newblock Model based training, detection and pose estimation of texture-less
  3d objects in heavily cluttered scenes.
\newblock In \emph{Asian conference on computer vision}, pages 548--562.
  Springer, 2012.

\bibitem[Hinterstoisser et~al.(2018)Hinterstoisser, Lepetit, Wohlhart, and
  Konolige]{hinterstoisser2018pre}
S.~Hinterstoisser, V.~Lepetit, P.~Wohlhart, and K.~Konolige.
\newblock On pre-trained image features and synthetic images for deep learning.
\newblock In \emph{Proceedings of the European Conference on Computer Vision
  (ECCV)}, pages 0--0, 2018.

\bibitem[Hinterstoisser et~al.(2019)Hinterstoisser, Pauly, Heibel, Martina, and
  Bokeloh]{hinterstoisser2019annotation}
S.~Hinterstoisser, O.~Pauly, H.~Heibel, M.~Martina, and M.~Bokeloh.
\newblock An annotation saved is an annotation earned: Using fully synthetic
  training for object detection.
\newblock In \emph{Proceedings of the IEEE International Conference on Computer
  Vision Workshops}, pages 0--0, 2019.

\bibitem[Huang et~al.(2018)Huang, Liu, Belongie, and
  Kautz]{huang2018multimodal}
X.~Huang, M.-Y. Liu, S.~Belongie, and J.~Kautz.
\newblock Multimodal unsupervised image-to-image translation.
\newblock In \emph{Proceedings of the European Conference on Computer Vision
  (ECCV)}, pages 172--189, 2018.

\bibitem[Isola et~al.(2017)Isola, Zhu, Zhou, and Efros]{isola2017image}
P.~Isola, J.-Y. Zhu, T.~Zhou, and A.~A. Efros.
\newblock Image-to-image translation with conditional adversarial networks.
\newblock In \emph{Proceedings of the IEEE conference on computer vision and
  pattern recognition}, pages 1125--1134, 2017.

\bibitem[Karras et~al.(2018)Karras, Aila, Laine, and
  Lehtinen]{karras2018progressive}
T.~Karras, T.~Aila, S.~Laine, and J.~Lehtinen.
\newblock Progressive growing of {GAN}s for improved quality, stability, and
  variation.
\newblock In \emph{International Conference on Learning Representations}, 2018.

\bibitem[Karras et~al.(2019)Karras, Laine, and Aila]{karras2019style}
T.~Karras, S.~Laine, and T.~Aila.
\newblock A style-based generator architecture for generative adversarial
  networks.
\newblock In \emph{Proceedings of the IEEE Conference on Computer Vision and
  Pattern Recognition}, pages 4401--4410, 2019.

\bibitem[Kehl et~al.(2017)Kehl, Manhardt, Tombari, Ilic, and
  Navab]{kehl2017ssd}
W.~Kehl, F.~Manhardt, F.~Tombari, S.~Ilic, and N.~Navab.
\newblock Ssd-6d: Making rgb-based 3d detection and 6d pose estimation great
  again.
\newblock In \emph{Proceedings of the International Conference on Computer
  Vision (ICCV 2017), Venice, Italy}, pages 22--29, 2017.

\bibitem[Mahmood et~al.(2018)Mahmood, Chen, and Durr]{mahmood2018unsupervised}
F.~Mahmood, R.~Chen, and N.~J. Durr.
\newblock Unsupervised reverse domain adaptation for synthetic medical images
  via adversarial training.
\newblock \emph{IEEE transactions on medical imaging}, 37\penalty0
  (12):\penalty0 2572--2581, 2018.

\bibitem[Oquab et~al.(2014)Oquab, Bottou, Laptev, and Sivic]{oquab2014learning}
M.~Oquab, L.~Bottou, I.~Laptev, and J.~Sivic.
\newblock Learning and transferring mid-level image representations using
  convolutional neural networks.
\newblock In \emph{Proceedings of the IEEE conference on computer vision and
  pattern recognition}, pages 1717--1724, 2014.

\bibitem[Prakash et~al.(2019)Prakash, Boochoon, Brophy, Acuna, Cameracci,
  State, Shapira, and Birchfield]{prakash2019structured}
A.~Prakash, S.~Boochoon, M.~Brophy, D.~Acuna, E.~Cameracci, G.~State,
  O.~Shapira, and S.~Birchfield.
\newblock Structured domain randomization: Bridging the reality gap by
  context-aware synthetic data.
\newblock In \emph{2019 International Conference on Robotics and Automation
  (ICRA)}, pages 7249--7255. IEEE, 2019.

\bibitem[Rambach et~al.(2018)Rambach, Deng, Pagani, and
  Stricker]{rambach2018learning}
J.~Rambach, C.~Deng, A.~Pagani, and D.~Stricker.
\newblock Learning 6dof object poses from synthetic single channel images.
\newblock In \emph{2018 IEEE International Symposium on Mixed and Augmented
  Reality Adjunct (ISMAR-Adjunct)}, pages 164--169. IEEE, 2018.

\bibitem[Rojtberg and Kuijper(2019)]{rojtberg2019real}
P.~Rojtberg and A.~Kuijper.
\newblock Real-time texturing for 6d object instance detection from rgb images.
\newblock In \emph{2019 IEEE International Symposium on Mixed and Augmented
  Reality Adjunct (ISMAR-Adjunct)}, pages 295--300. IEEE, 2019.

\bibitem[Tekin et~al.(2018)Tekin, Sinha, and Fua]{tekin2018real}
B.~Tekin, S.~N. Sinha, and P.~Fua.
\newblock Real-time seamless single shot 6d object pose prediction.
\newblock In \emph{Proceedings of the IEEE Conference on Computer Vision and
  Pattern Recognition}, pages 292--301, 2018.

\bibitem[Tobin et~al.(2017)Tobin, Fong, Ray, Schneider, Zaremba, and
  Abbeel]{tobin2017domain}
J.~Tobin, R.~Fong, A.~Ray, J.~Schneider, W.~Zaremba, and P.~Abbeel.
\newblock Domain randomization for transferring deep neural networks from
  simulation to the real world.
\newblock In \emph{2017 IEEE/RSJ international conference on intelligent robots
  and systems (IROS)}, pages 23--30. IEEE, 2017.

\bibitem[Torralba and Efros(2011)]{torralba2011unbiased}
A.~Torralba and A.~A. Efros.
\newblock Unbiased look at dataset bias.
\newblock In \emph{CVPR 2011}, pages 1521--1528. IEEE, 2011.

\bibitem[Tremblay et~al.(2018{\natexlab{a}})Tremblay, Prakash, Acuna, Brophy,
  Jampani, Anil, To, Cameracci, Boochoon, and Birchfield]{tremblay2018training}
J.~Tremblay, A.~Prakash, D.~Acuna, M.~Brophy, V.~Jampani, C.~Anil, T.~To,
  E.~Cameracci, S.~Boochoon, and S.~Birchfield.
\newblock Training deep networks with synthetic data: Bridging the reality gap
  by domain randomization.
\newblock In \emph{Proceedings of the IEEE Conference on Computer Vision and
  Pattern Recognition Workshops}, pages 969--977, 2018{\natexlab{a}}.

\bibitem[Tremblay et~al.(2018{\natexlab{b}})Tremblay, To, Sundaralingam, Xiang,
  Fox, and Birchfield]{tremblay2018deep}
J.~Tremblay, T.~To, B.~Sundaralingam, Y.~Xiang, D.~Fox, and S.~Birchfield.
\newblock Deep object pose estimation for semantic robotic grasping of
  household objects.
\newblock In \emph{Conference on Robot Learning}, pages 306--316,
  2018{\natexlab{b}}.

\bibitem[Tsirikoglou et~al.(2017)Tsirikoglou, Kronander, Wrenninge, and
  Unger]{tsirikoglou2017procedural}
A.~Tsirikoglou, J.~Kronander, M.~Wrenninge, and J.~Unger.
\newblock Procedural modeling and physically based rendering for synthetic data
  generation in automotive applications.
\newblock \emph{arXiv preprint arXiv:1710.06270}, 2017.

\bibitem[Wang et~al.(2018)Wang, Liu, Zhu, Tao, Kautz, and
  Catanzaro]{wang2018high}
T.-C. Wang, M.-Y. Liu, J.-Y. Zhu, A.~Tao, J.~Kautz, and B.~Catanzaro.
\newblock High-resolution image synthesis and semantic manipulation with
  conditional gans.
\newblock In \emph{Proceedings of the IEEE conference on computer vision and
  pattern recognition}, pages 8798--8807, 2018.

\bibitem[Xiang et~al.(2018)Xiang, Schmidt, Narayanan, and Fox]{Xiang-RSS-18}
Y.~Xiang, T.~Schmidt, V.~Narayanan, and D.~Fox.
\newblock Posecnn: A convolutional neural network for 6d object pose estimation
  in cluttered scenes.
\newblock In \emph{Proceedings of Robotics: Science and Systems}, Pittsburgh,
  Pennsylvania, June 2018.
\newblock \doi{10.15607/RSS.2018.XIV.019}.

\bibitem[Yu et~al.(2015)Yu, Seff, Zhang, Song, Funkhouser, and
  Xiao]{yu2015lsun}
F.~Yu, A.~Seff, Y.~Zhang, S.~Song, T.~Funkhouser, and J.~Xiao.
\newblock Lsun: Construction of a large-scale image dataset using deep learning
  with humans in the loop.
\newblock \emph{arXiv preprint arXiv:1506.03365}, 2015.

\bibitem[Zakharov et~al.(2019)Zakharov, Kehl, and
  Ilic]{zakharov2019deceptionnet}
S.~Zakharov, W.~Kehl, and S.~Ilic.
\newblock Deceptionnet: Network-driven domain randomization.
\newblock In \emph{Proceedings of the IEEE International Conference on Computer
  Vision}, pages 532--541, 2019.

\bibitem[Zhu et~al.(2017)Zhu, Park, Isola, and Efros]{CycleGAN2017}
J.-Y. Zhu, T.~Park, P.~Isola, and A.~A. Efros.
\newblock Unpaired image-to-image translation using cycle-consistent
  adversarial networks.
\newblock In \emph{Computer Vision (ICCV), 2017 IEEE International Conference
  on}, 2017.

\end{thebibliography}

\end{document}